\documentclass{article}

\usepackage{microtype}
\usepackage{graphicx}
\usepackage{subfigure}
\usepackage{booktabs} % for professional tables
\usepackage{nth}

\usepackage[colorlinks=true,linkcolor=blue,citecolor=blue]{hyperref}

\usepackage[accepted]{icml2020}

\usepackage[utf8]{inputenc} % allow utf-8 input
\usepackage[T1]{fontenc}    % use 8-bit T1 fonts
\usepackage{url}            % simple URL typesetting
\usepackage{booktabs}       % professional-quality tables
\usepackage{amsfonts}       % blackboard math symbols
\usepackage{nicefrac}       % compact symbols for 1/2, etc.
\usepackage{amsmath}
\usepackage{amsthm}
\usepackage{graphicx}
\usepackage{multirow}
\usepackage{url}            % simple URL typesetting
\usepackage{nicefrac}       % compact symbols for 1/2, etc.
\usepackage{microtype}      % microtypography
\usepackage{amssymb}
\usepackage{xcolor,colortbl}
\usepackage{multirow}
\usepackage{tabu}
\usepackage{wrapfig}
\usepackage{blindtext}
\usepackage{inconsolata}

\definecolor{aurometalsaurus}{rgb}{0.43, 0.5, 0.5}
\definecolor{unigreen}{rgb}{0.0,0.5961,0.3765}

\theoremstyle{definition}
\newtheorem{defi}{Definition}

\theoremstyle{plain}

\newtheorem{lem}{Lemma}

\theoremstyle{definition}
\newtheorem{rem}{Remark}

\newcommand{\ph}{\texttt{ph}}
\newcommand{\wrt}{{w.r.t.}}
\newcommand{\ie}{{i.e.}}
\newcommand{\eg}{{e.g.}}

\DeclareMathOperator{\lab}{\textup{\text{lab}}}

 \graphicspath{{figs/}}
\newcommand{\bbB}{\mathbb{B}}
\newcommand{\bbG}{\mathbb{G}}

\newcommand{\bbK}{\mathbb{K}}

\newcommand{\R}{\mathbb{R}}

\newcommand{\bbV}{\mathbb{V}}

\newcommand{\bbY}{\mathbb{Y}}
\newcommand{\Z}{\mathbb{Z}}

\newcommand{\mcB}{\mathcal{B}}

\newcommand{\mcP}{\mathcal{P}}

\newcommand{\mcV}{\mathcal{V}}

\newcommand\restr[2]{{% we make the whole thing an ordinary symbol
  \left.\kern-\nulldelimiterspace % automatically resize the bar with \right
  #1 % the function
  \vphantom{\big|} % pretend it's a little taller at normal size
  \right|_{#2} % this is the delimiter
  }}
\DeclareMathOperator{\im}{im}
\DeclareMathOperator{\rank}{rank}

\definecolor{sky0}{HTML}{375E97}
\definecolor{sky1}{HTML}{5276AA}
\definecolor{sky2}{HTML}{1E4987}
\definecolor{sunset0}{HTML}{FB6542}
\definecolor{sunset1}{HTML}{FF8366}
\definecolor{sunset2}{HTML}{EF3C11}
\definecolor{sunflower0}{HTML}{FFBB00}
\definecolor{sunflower1}{HTML}{FFCA39}
\definecolor{sunflower2}{HTML}{C69100}

\definecolor{grass0}{HTML}{3F681C}
\definecolor{grass1}{HTML}{578134}
\definecolor{grass2}{HTML}{264809}

\icmltitlerunning{Graph Filtration Learning}

\pagecolor{white}
\begin{document}

\twocolumn[
    \icmltitle{Graph Filtration Learning}



    \icmlsetsymbol{equal}{*}

    \begin{icmlauthorlist}
    \icmlauthor{Christoph D. Hofer}{sbg}
    \icmlauthor{Florian Graf}{sbg}
    \icmlauthor{Bastian Rieck}{eth}
    \icmlauthor{Marc Niethammer}{unc}
    \icmlauthor{Roland Kwitt}{sbg}    
    \end{icmlauthorlist}
    
    \icmlaffiliation{sbg}{Department of Computer Science, University of Salzburg, Austria}
    \icmlaffiliation{unc}{UNC Chapel Hill}
    \icmlaffiliation{eth}{Department of Biosystems Science and Engineering, ETH Zurich, Switzerland}

    \icmlcorrespondingauthor{Christoph D. Hofer}{\texttt{chr.dav.hofer@gmail.com}}

    \icmlkeywords{Machine Learning, Graphs, Persistent Homology}

    \vskip 0.3in
]



\printAffiliationsAndNotice{}  % otherwise use the standard text.

\begin{abstract}
    % auto-ignore
% !TEX root = ./graph_filtration_learning.tex

We propose an approach to learning with graph-structured data in the 
problem domain of graph classification. In particular, we present a novel 
type of \emph{readout} operation to aggregate node features into a
graph-level representation. To this end, we leverage persistent homology
computed via a real-valued, learnable, filter function. 
We establish the theoretical foundation for differentiating through the 
persistent homology computation. Empirically, we show that this type of
readout operation compares favorably to previous techniques, 
especially when the graph connectivity structure is informative for 
the learning problem.

\end{abstract}

% !TEX root = ./graph_filtration_learning.tex

\section{Introduction}
\label{section:introduction}
We consider the task of learning a function from the space of (finite) undirected graphs, $\bbG$, to a (discrete/continuous) target domain $\bbY$. Additionally, graphs might have 
discrete, or continuous attributes attached to each node. Prominent examples for this class of \emph{learning} problem appear in the context of classifying molecule structures, chemical compounds or social networks.
\begin{figure}[h!]
    \centering{
    \includegraphics[width=\columnwidth]{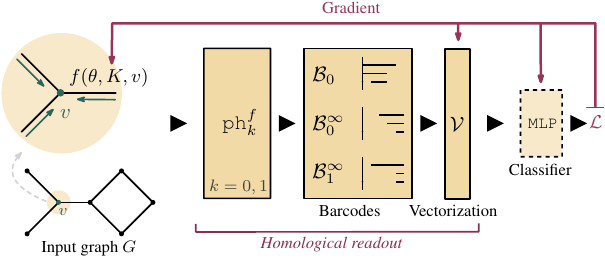}}
    \vspace{-4pt}
    \caption{\label{fig:intro}
    Overview of the proposed \emph{homological readout}. Given a graph/simplicial complex, we use a vertex-based graph functional $f$ to assign a real-valued score to  each node.
    A practical choice is to implement $f$ as a $\texttt{GNN}$, with one level of message passing. 
    We then compute persistence barcodes, $\mcB_k$, using the filtration induced by $f$.
    Finally, barcodes are fed through a vectorization scheme $\mcV$ and passed to a classifier 
    (e.g., an \texttt{MLP}). Our approach allows passing a learning signal \emph{through} the persistent homology computation, allowing to optimize $f$ for the classification task.}
\end{figure}

A substantial amount of research has been devoted to developing techniques for supervised learning with graph-structured data, ranging from kernel-based methods \cite{Shervashidze09a,Shervashidze11a,Feragen13a,Kriege16a}, to more recent approaches based on graph neural networks (GNN) \cite{Scarselli09a,Hamilton17b,Zhang18a,Morris19a,Xu19a,Ying18a}. 
Most of the latter works use an iterative message passing scheme \cite{Gilmer17a} to learn node representations, followed by a graph-level pooling operation that aggregates node-level features. This aggregation step is typically referred 
to as a \emph{readout} operation. While research has mostly focused on variants of the message passing function, the readout step may have a significant impact, as it aims to capture properties of the entire graph. 
%As remarked in \cite{Ying18a}, commonly used readouts, such as sum or average pooling can
%be suboptimal, as they ignore the connectivity structure of a graph. 
Importantly, both simple and more refined readout operations,
such as summation, differentiable pooling \cite{Ying18a}, or sort pooling \cite{Zhang18b}, 
are inherently coupled to the amount of information carried over via 
multiple rounds of message passing. Hence, architectural GNN choices 
are typically guided by dataset characteristics, e.g., requiring to tune 
the number of message passing rounds to the expected size of graphs.

\textbf{Contribution.}
We propose a \emph{homological} readout operation that captures the full 
global structure of a graph, while relying only on node representations 
learned from immediate neighbors. 
This not only alleviates the aforementioned design challenge, but potentially 
offers additional discriminative information. Similar to previous
works, we consider a graph, $G$, as a simplicial complex, $K$,   
and use persistent homology \cite{Edelsbrunner2010} to capture  
homological changes that occur when constructing the graph one 
part at a time (\ie, revealing changes in the number of connected 
components or loops). As this hinges on an ordering of the parts, prior
works rely on a suitable \emph{filter function} $f:K \rightarrow \R$
to establish this ordering. \emph{Our idea differs in that we 
learn the filter function (end-to-end), as opposed to defining it a-priori)
}.

%\textbf{Contributions.} In this work, we go one step further and propose to \emph{learn} the filter 
%function in an end-to-end manner, essentially requiring the capacility to 
%backpropagate a learning signal through the persistent homology computation. 
%Combined with previous work on (1) learning node representations and 
%on (2) learning with topological summaries, this offers a novel type of \emph{readout}.
%operation. Experiments on various benchmark datasets indicate that even by using a 
%simple GNN with one level of message passing, our readout operator captures sufficient
%information to achieve state-of-the-art performance and is superior to other readout 
%techniques in the same setup. 
%Additionally, as the learnt filter $f$ maps to 
%$\R$, we can easily color nodes, offering additional qualitative information.

% auto-ignore
% !TEX root = ./graph_filtration_learning.tex

\section{Related work}
\label{section:related_work}

\textbf{Graph neural networks.} Most work on neural network based approaches to learning with graph-structured
data focuses on learning informative \emph{node embeddings} to solve tasks such as 
link prediction \cite{Schuett17a}, node classification \cite{Hamilton17b}, or classifying 
entire graphs. Many of these approaches, 
including \cite{Scarselli09a, Duvenaud15a,Li16a,Battaglia16a,Kearns16a,Morris19a}, can be formulated
as a \emph{message passing} scheme \cite{Gilmer17a,Xu18a} where features of graph 
nodes are passed to immediate neighbors via a differentiable message passing function; this operation proceeds over multiple iterations with each iteration parameterized by a neural network. 
%Earlier work \cite{} even iterates until convergence (possibly under certain 
%conditions on the message passing function). 
Aspects distinguishing these approaches include (1) the 
particular realization of the message passing function, (2) the way information is aggregated at the nodes, and (3) whether edge features are included. Due to the algorithmic similarity of iterative message passing and aggregation to the Weisfeiler-Lehman (WL) graph 
isomorphism test \cite{Weisfeiler68a}, several works \cite{Xu19a,Morris19a} have recently studied this connection 
and established a theoretical underpinning for analyzing properties of 
GNN variants in the context of the WL test.

\textbf{Readout operations.} 
With few exceptions, surprisingly little effort has been devoted to the so called 
\emph{readout} operation, i.e., a function that aggregates node features into a global graph
representation and allows making predictions for an entire graph.
Common strategies include summation \cite{Duvenaud15a}, averaging, or passing
node features through a network operating on sets \cite{Li16a, Gilmer17a}. As 
pointed out in \cite{Ying18a}, this effectively ignores the often complex 
global hierarchical structure of graphs. To mitigate this issue, \cite{Ying18a}  
proposed a differentiable pooling operation that interleaves each message passing 
iteration and successively coarsens the graph. A different pooling scheme is 
proposed in \cite{Zhang18b}, relying on appropriately sorting node features
obtained from each message passing iteration. Both pooling mechanisms are generic 
and show improvements on multiple benchmarks. Yet, gains inherently
depend on multiple rounds of message passing, as the global structure is successively 
captured during this process. In our alternative approach, \emph{global} structural 
information is captured even initially.
It only hinges on attaching a real value to each node which is learnably implemented via a GNN in one round of message passing.

\textbf{Persistent homology \& graphs.} Notably, analyzing graphs via persistent 
homology is not new, with several works showing 
promising results on graph classification \cite{Hofer17a,Carriere19a,Rieck19a,Zhao19}. 
So far, however, persistent homology is used in a \emph{passive} 
manner, meaning that the function $f$ mapping simplices to $\mathbb{R}$ is \emph{fixed} 
and not informed by the learning task.
Essentially, this degrades persistent homology
to a feature extraction step, where the obtained topological summaries are 
fed through a vectorization scheme and then passed to a classifier.
The success of these methods inherently hinges on the choice of
the a-priori defined function $f$, \eg, the node degree function in \cite{Hofer17a} 
or a heat kernel function in \cite{Carriere19a}. \emph{The difference
to our approach is that backpropagating the learning signal stops at
the persistent homology computation; in our case, the signal 
is passed through, allowing to adjust $f$ during learning}.

% auto-ignore
% !TEX root = ./graph_filtration_learning.tex

\section{Background}
\label{section:background}

We are interested in describing graphs in terms of their topological features, such as connected components and cycles. 
Persistent homology, a method from computational topology, makes it possible to compute these features efficiently. 
Next, we provide a concise introduction to the necessary concepts of persistent homology and refer the reader to \cite{Hatcher2002, Edelsbrunner2010} for details. 

\textbf{Homology.}
The key concept of homology theory is to study properties of an object $X$, such as a graph, by means of (commutative) algebra. Specifically, we assign to $X$ a sequence of groups/modules $C_0,\ C_1, \ldots$ which are connected by homomorphisms $\partial_{k+1}: C_{k+1}\rightarrow C_{k}$ such that 
$\im \partial_{k+1} \subseteq \ker \partial_{k}$. 
A structure of this form is called a \emph{chain complex} and by studying its homology groups 
\begin{equation}
H_k = \ker \partial_{k} / \im \partial_{k+1}
\label{eq:homology_group}
\end{equation}
we can derive (homological) properties of $X$. 
The original motivation for homology is to analyze topological spaces. 
In that case, the ranks of homology groups yield directly interpretable properties, \eg, $\rank(H_0)$ reflects the number of connected components and $\rank(H_1)$ the number of loops. 

A prominent example of a homology theory is \emph{simplicial homology}.
A simplicial complex, $K$, over the vertex domain $\bbV$ is a set of non-empty (finite) subsets of $\bbV$ that is closed under the operation of taking non-empty subsets and does not contain $\emptyset$. 
Formally, this means $K \subset \mcP(\bbV)$ with $\sigma \in K \Rightarrow 1 \leq |\sigma| < \infty$ and
$\tau \subseteq \sigma \in K \Rightarrow \tau \in K$.
We call $\sigma \in K$ a $k-$simplex iff $\dim(\sigma) = |\sigma| - 1 = k$; correspondingly 
$\dim(K)=\max_{\sigma \in K}\dim(\sigma)$ and 
we set $K_k = \{\sigma \in K: \dim(\sigma) = k\}$.
%and call $\cup_{k' \leq k}K_{k'}$ its $k$\emph{-skeleton}. 
Further, let $C_k(K)$ be the vector space generated by $K_k$ over $\Z/2\Z$\footnote{Simplicial homology is not specific 
to $\Z/2\Z$, but it's a typical choice, since it allows us to interpret $k$-chains as sets
of $n$-simplices.} and define 
\[\partial_k:K_k \rightarrow C_{k-1}(K) 
\quad 
\sigma \mapsto \sum\limits_{\substack{\tau: \tau \subset \sigma \\ \dim(\tau) = k-1}} \tau
\enspace.
\]
In other words, $\sigma$ is mapped to the formal sum of its $(k-1)$-dim. faces, \ie, 
its subsets of cardinality $k$. 
The linear extension to $C_k(K)$ of  this mapping defines the $k$-th \emph{boundary operator} 
\begin{align*}
\partial_k: C_k(K) &\to C_{k-1}(K) 
\\
\sum\limits_{i=1}^n \sigma_i &\mapsto \sum\limits_{i=1}^n \partial_{k}(\sigma_i)\enspace.
\end{align*}
Using $\partial_k$, we obtain the corresponding $k$-th homology group, $H_k$, as in Eq. \eqref{eq:homology_group}.

\begin{figure*}
\begin{center}
\includegraphics[width=0.90\textwidth]{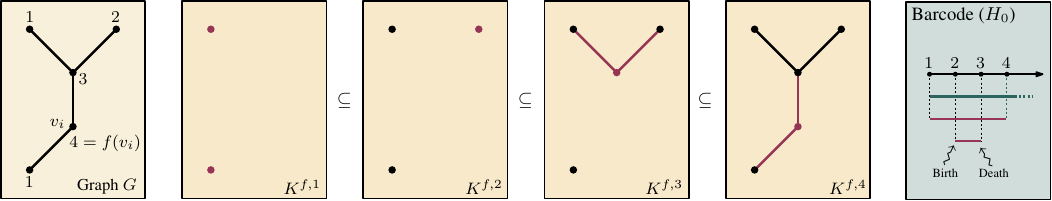}
\end{center}
\vspace{-5pt}
\caption{\label{fig:gfl}Illustration of $0$-dimensional persistent homology on a graph $G$, via a node-based \emph{filter} function $f$. The
barcode (shown right), computed by persistent homology is $\mathcal{B}_0=\{(1,4),(2,3)\}$. If we consider \emph{essential} topological
features, we further get $\mathcal{B}_0^{\infty}=\{1\}$.}
\end{figure*}

{\bf Graphs are simplicial complexes.} 
In fact, we can directly interpret a graph $G$ as an $1$-dimensional simplicial complex whose vertices ($0$-simplices) are the graphs nodes and whose edges ($1$-simplices) are the graphs edges. 
In this setting, only the \nth{1} boundary operator is non trivial and simply maps an edge to the formal sum of its defining nodes. 
Via homology we can then compute the graph's $0$ and $1$ dimensional topological features, \ie, connected components and cycles. 
However, as mentioned earlier, this only gives a rather coarse summary of the graph's topological properties, raising the demand for a \emph{refinement} of homology. 

\textbf{Persistent homology.}
For consistency with the relevant literature, we introduce the idea of refining homology to \emph{persistent} homology in terms of simplicial complexes, but recall that graphs \emph{are} $1$-dimensional simplicial complexes. 

Now, let $(K^i)_{i=0}^{m}$ be a sequence of simplicial complexes such that $\emptyset=K^0 \subseteq K^1 \subseteq \dots \subseteq K^m = K$. Then, $(K^i)_{i=0}^{m}$ is called a \emph{filtration} of $K$. 
Using the extra information provided by the filtration of $K$, we obtain a sequence of chain complexes 
%(\emph{left}), 
%\begin{center}
%\includegraphics[width=\columnwidth]{cc}
%\end{center}
where $C_k^i = C_k(K_k^i)$
and $\iota$ denotes the inclusion.
This leads to the concept of \emph{persistent homology groups}, \ie,
\[
H_k^{i,j} = \ker \partial_{k}^i / (\im \partial_{k+1}^j \cap \ker \partial_k^i) \quad\text{for}\quad 1 \leq i \leq j \leq m\enspace.
\]
The ranks, $\beta_k^{i,j} = \rank H_k^{i,j}$, of these homology groups (i.e., the \emph{$k$-th 
persistent Betti numbers}), capture the number of homological features of dimensionality 
$k$ (e.g., connected components for $k=0$, loops for $k=1$, etc.) that \emph{persist} from $i$ to (at least) $j$.
According to the Fundamental Lemma of Persistent Homology \cite{Edelsbrunner2010},
the quantities
\begin{equation}
\mu_k^{i,j} = (\beta_k^{i, j-1} - \beta_k^{i,j}) - (\beta_k^{i-1, j-1} - \beta_k^{i-1, j})
\label{eqn:munij}
\end{equation}
for $1 \leq i < j \leq m$
encode all the information about the persistent Betti numbers of dimension $k$.

\textbf{Persistence barcodes.} In principle, introducing persistence barcodes 
only requires a filtration as defined above. In our setting, we
define a filtration of $K$ via a vertex filter function 
$f:\bbV \rightarrow \R$. In particular, given the sorted sequence of filter values
$a_1 < \cdots < a_m$, \ie, $a_i \in \{f(v): \{v\} \in K_0\}$, 
we define the filtration \wrt $f$ as
\begin{equation}
\label{eq:filtration}
\begin{split}
K^{f,0} & = \emptyset \\
K^{f,i} & = \{\sigma \in K: \max\limits_{v \in \sigma} f(v) \leq a_i\} \\
\end{split}
\end{equation}
for $ 1 \leq i \leq m$. Intuitively, this means that the filter function is defined on the vertices of $K$ and lifted to higher dimensional 
simplices in $K$ by maximum aggregation. This enables us to consider a sub levelset filtration of $K$. 
% A way to obtain a filtration of $K$ is to define a a function 
% $f:\bbV \rightarrow \R$, and lift it to $K_k$, $k \geq 1$, by maximum aggregation, \ie,  
% \[
% f(\sigma) = \max \{f(v): v \in \sigma\}
% \]
% and consider the set tower of sublevel sets as a filtration of $K$. 
% To be precise, given $m = |f(K_0)|$, we obtain $(K^{f, i})_{i=0}^m$ by setting 
% \begin{equation}
% $K^{f, 0} =\emptyset$ and $K^{f, i} = f^{-1}((-\infty, a_i])$ 
% \end{equation}
% for $1 \leq i \leq m$, where $a_1 < \cdots < a_m$
% is the sorted sequence of filtration values, \ie, $f(K_0)$.

Then, for a given filtration of $K$ and $0 \leq k \leq \dim(K)$, we can construct a multiset by inserting the point $(a_i, a_j)$, $1 \leq i<j \leq m$, with multiplicity $\mu_k^{i,j}$. 
This effectively encodes the $k$-dimensional persistent homology of $K$ \wrt~the given filtration. This representation is called a \emph{persistence barcode}, $\mcB_k$.
% , for $1 \leq i<j \leq m$, the point $(a_i, a_j)$ is inserted with multiplicity $\mu_k^{i,j}$, we effectively encode the persistent homology of dimension $k$ w.r.t. the filtration induces by $f$, see Eq. \eqref{eq:filtration}.
% We call this representation, yielded by persistent homology, a \emph{persistence barcode}.
% 
% \begin{defi}[Persistence barcode]
%     Let $\Delta = \{x \in \R^2_{\Delta}: \multiplicity(x) = \infty\}$ be the multiset of the 
%     diagonal $\R^2_{\Delta} = \{(x_0, x_1) \in \R^2: x_0 = x_1\}$, where $\multiplicity$ denotes the multiplicity function and let 
%     $\R^2_{\star} = \{(x_0, x_1) \in \R^2: x_1 > x_0\}$. 
%     A persistence diagram, $\mcD$, is a multiset of the form
%     \[
%     \mcD = \{x: x \in \R^2_{\star}\} \cup \Delta\enspace.
%     \]
%     We denote by $\bbD$ the set of all persistence diagrams of the 
%     form $|\mcD \setminus \Delta|<\infty\enspace.$
%     \label{defn:pd}
% \end{defi}
% 
Hence, for a given complex $K$ of dimension $\dim(K)$ and a filter function $f$ (of the discussed form),  
we can interpret $k$-dimensional persistent homology as a mapping of simplicial complexes, defined by
\begin{equation}
\label{eq:ph_as_mapping}
  \ph_k^f(K) = \mcB_k \quad 0 \leq k \leq  \text{dim}(K)\enspace.
\end{equation}
%where $\mcB_k$ is the barcode of dimension $k$.
\begin{rem}
By setting 
 \[
 \mu_k^{i, \infty} = \beta_n^{i, m} - \beta_n^{i-1, m}\enspace
 \]
 we extend Eq.~\eqref{eqn:munij} to features which never disappear, also referred to as \emph{essential}. 
  If we use this extension, Eq.~\eqref{eq:ph_as_mapping} yields an additional barcode, denoted as $\mcB_k^{\infty}$, per dimension. 
  For practical reasons, the points in $\mcB_k^{\infty}$ are just the birth-time, as all death-times equal to $\infty$ and thus are omitted. 
 \end{rem}

{\bf Persistent homology on graphs.} 
In the setting of graph classification, we can think of a filtration as a growth process of a weighted graph. 
As the graph grows, we track homological changes and gather this information in its 0- and 1-dimensional persistence barcodes. An illustration of $0$-dimensional persistent homology for
a toy graph example is shown in Fig.~\ref{fig:gfl}.

% auto-ignore
% !TEX root = ./graph_filtration_learning.tex

\section{Filtration learning}
\label{section:filtration_learning}

Having introduced the required terminology, we now focus on the main
contribution of this paper, i.e., how to \emph{learn} an appropriate
filter function. To the best of our knowledge, our paper constitutes the
first approach to this problem. While different filter functions for 
graphs have been proposed and used in the past, \eg, based on
degrees~\citep{Hofer17a}, cliques~\citep{Petri13}, multiset
distances~\citep{Rieck19}, or the Jaccard distance~\citep{Zhao19},
prior work typically treats the filter function as fixed. In this setting,
the filter function needs to be selected via, \eg, cross-validation, 
which is cumbersome. 
Instead, we show that it is possible to learn a filter function \emph{end-to-end}.

This endeavor hinges on the capability of backpropagating gradients
through the persistent homology computation. While this has previously been 
done in \cite{Chen19a} or \cite{Hofer19b}, their results are 
not directly applicable here, due to the special problem settings 
considered in these works. In \cite{Chen19a}, the authors regularize 
decision boundaries of classifiers, in \cite{Hofer19b} the authors
impose certain topological properties on representations learned 
by an autoencoder. This warrants a detailed analysis in the context
of graphs.

We start by recalling that the computation of sublevel set
persistent homology, cf.~Eq.~\eqref{eq:filtration}, depends on two
arguments: (1) the complex $K$ and (2) the filter function $f$ which
determines the order of the simplices in the filtration of $K$.
As $K$ is of discrete nature, it is clear that $K$ cannot be
subject to gradient-based optimization. However, assume that the filter
$f$ has a differentiable dependence on a real-valued parameter $\theta$.
In this case, the persistent homology of the sublevel set filtration of $K$
\emph{also} depends on $\theta$, raising the following question: \emph{Is the mapping differentiable in $\theta$?}

%
%
%
%First, note that the computation of sublevel set persistent homology, cf.~Eq.~\eqref{eq:filtration}, is dependent on two arguments: (1) the complex $K$ and (2) the filter function $f$ which determines the order of the simplices in the filtration of $K$.
%As $K$ is inherently of discrete nature, it is clear that $K$ can not be subject to gradient based optimization. However, assume that the filter $f$ has a differentiable dependence on a real-valued parameter $\theta$. Then, also the persistent homology of the sub levelset filtration of $K$ is dependent on $\theta$. 
%%Practically this means that for a fixed $K$ we can interpret $(K, \theta) \mapsto \ph_k^f(K, \theta)$ as a mapping %dependent only on $\theta$. 
%This immediately raises the following question: \emph{Is the mapping differentiable in $\theta$?}\\
% 
\textbf{Notation.} If any symbol introduced in the context of persistent homology is dependent on $\theta$, we interpret it as function of $\theta$ and attach ``$(\theta)$'' to this symbol. If the dependence on the simplicial complex $K$ is irrelevant to the current context, we omit it for brevity. 
For example, we write $\mu_k^{i, j}(\theta)$ for the multiplicity of barcode points.

Next, we concretize the idea of a learnable vertex filter. 
 % 
% auto-ignore
% !TEX root = ./graph_filtration_learning.tex

\vskip1ex
\begin{defi}[Learnable vertex filter function]
\label{defi:learnable_filter}
Let $\bbV$ be a vertex domain, $\bbK$ the set of possible simplicial complexes over $\bbV$
and let
\[
f: \R \times \bbK \times \bbV \rightarrow \R \quad\quad (\theta, K, v) \mapsto f(\theta,K, v)
\]
be differentiable in $\theta$ for $K \in \bbK, v \in \bbV$. Then, we call $f$ a \emph{learnable vertex filter function} with parameter $\theta$. 
\end{defi}
The assumption that $f$ is dependent on a \emph{single} parameter is solely dedicated to simplify the following theoretical part. The derived results immediately generalize to the case	 $f: \R^n \times \bbK \times \bbV$ where $n$ is the number of parameters and we will drop this assumption later on. 

% 
% As previously stated, Eq. \eqref{eq:filtration}, $f$ allows us to define a valid filtration of $K$.  
% For notational convenience we will write $\ph_k^f(\theta, K)$ for the persistent homology computation \wrt~ this filtration. 

By Eq.~\eqref{eq:ph_as_mapping},  $(\theta, K) \mapsto \ph_k^f(\theta, K)$ is a mapping to the space of persistence barcodes $\bbB$ (or $\bbB^2$ if essential barcodes are included). As $\bbB$ has no natural linear structure, we consider differentiability in combination with a coordinatization strategy $\mcV: \bbB \rightarrow \R$. This allows studying the map 
$$( \theta, K) \mapsto \mcV\big(\ph_k^f(\theta, K)\big)$$ 
for which differentiability is defined. In fact, we can rely on prior works on learnable
vectorization schemes for persistence barcodes \cite{Hofer19a, Carriere19a}. In the following definition of a barcode coordinate function, 
we adhere to \cite{Hofer19b}.
%Using persistence barcodes as input to machine learning algorithms has already found several applications~\cite{Bendich2016,Kwitt15a,Hofer17a,Adams17a,Carriere19a,Zhao19}. 
%Technically, the multiset nature of barcodes prevents the direct application of standard methods, such as SVMs or neural networks. 
%The two dominant strategies to mitigate this problem are: vectorization and kernel-based learning. As we need a mechanism to handle barcodes once the persistent homology is 
%computed via 
% 
% auto-ignore
% !TEX root = ./graph_filtration_learning.tex

\begin{defi}[Barcode coordinate function]
\label{defi:barcode_coordinate_function}
Let $s: \R^2 \rightarrow \R$ be a differentiable function that vanishes on the
diagonal of $\mathbb{R}^2$. Then 
\[
  \mcV: \bbB \rightarrow \R \quad \quad \mcB \mapsto \sum\limits_{(b, d) \in \mcB}s(b, d)
\]
is called \emph{barcode coordinate function}. 
\label{defi:barcode_coordinate_function}
\end{defi}
Intuitively, $\mcV$ maps a barcode $\mcB$ to a real value by aggregating the points in $\mcB$ via a weighted sum. 
%In fact, the (vectorization) input layer presented in \cite{Hofer17a} falls withing the family of mappings defined in Definition \ref{defi:barcode_coordinate_function}, 
Notably, the deep sets approach of \cite{Zaheer2017a} would also be a natural choice, however,
not specifically tailored to persistence barcodes. 
Upon using $d$ barcode coordinate functions of the form described in Definition~\ref{defi:barcode_coordinate_function}, we can effectively map a barcode into $\mathbb{R}^d$ and feed this representation through any differentiable layer downstream, e.g., implementing a 
classifier. We will discuss our particular choice of $s$ in \S\ref{section:experiments}. 

Next, we show that, under certain conditions, $\ph_k^f$ in combination with a suitable barcode coordinate function preserves differentiability. This is the \emph{key result} that will enable end-to-end learning of a filter function.
% 
% auto-ignore
% !TEX root = ./graph_filtration_learning.tex

\begin{lem}
\label{lemma:filter_differentiable}
Let $K$ be  a \emph{finite} simplicial complex with vertex set $V=\{v_1, \dots, v_n\}$, $f: \R \times \bbK \times \bbV \rightarrow \R$ be a learnable vertex filter function as in Definition \ref{defi:learnable_filter} and $\mcV$ a barcode coordinate function as in Definition \ref{defi:barcode_coordinate_function}.
If, for $\theta_0 \in \R$, it holds that the pairwise vertex filter values are distinct, \ie, 
\[
  f(\theta_0, K, v_i) \neq f(\theta_0,K,  v_j) 
  \quad \text{ for } \quad 
  1 \leq i < j \leq n
\]
then the mapping
\begin{equation}
  \theta \mapsto \mcV\big(\ph_k^{f}(\theta, K)\big)
\label{eqn:lemma:pairwise_filter_values_differentiable}
\end{equation}
is \emph{differentiable} at $\theta_0$. 
\end{lem}

For brevity, we only sketch the proof; the full version can be found in the supplementary material.

\begin{proof}[Sketch of proof]
The pairwise vertex filter values
$$Y = \{f(\theta_0, K, v_i)\}_{i=1}^n$$
are distinct
which implies that they are (\textbf{P1}) strictly ordered by ``$<$'' and (\textbf{P2}) $m = |Y| = n$.
A cornerstone for the actual persistence computation is the index permutation $\pi$  which sorts $Y$.
Now consider for some $h \in \R$ the set of vertex filter values for $\theta_0$ shifted by $h$, \ie,  $Y' = \{f(\theta_0 + h, K, v_i)\}_{i=1}^n$. 
Since $f$ is assumed to be differentiable, and therefore continuous, $Y'$ also satisfies (P1) and (P2) 
and $\pi$ sorts $Y'$, for $|h|$ sufficiently small.  
Importantly, this also implies 
\begin{equation}
  \big(K_i^f(\theta_0)\big)_{i=0}^n = \big(K_i^f(\theta_0 + h)\big)_{i=0}^n
\end{equation}
and thus
\begin{equation}
  \label{lemma:filter_differentiable:eqn:filt}
  \mu_k^{i, j}(\theta_0) =  \mu_k^{i, j}(\theta_0 + h)
  ~
  \text{for}
  ~
  1 \leq i < j \leq n  \enspace. 
\end{equation}
As a consequence, this allows deriving the equality 
\begin{small}
\begin{gather*}
\lim\limits_{|h| \rightarrow 0} 
\frac{\mcV(\ph_k^f(K,\theta_0)) - \mcV(\ph_k^f(K,\theta_0+h))}{h}  \\
=\\  
\sum\limits_{i < j}
  \mu_k^{i, j}(\theta_0) \cdot 
  \frac{\partial s\big(f(\theta,K,v_{\pi(i)}), f(\theta,K,v_{\pi(j)})\big)}{\partial \theta}(\theta_0)\enspace.
\end{gather*}
\end{small}

This concludes the proof, since the derivative within the summation on the right exists 
(as $f$ was assumed to be differentiable).
\end{proof}

\begin{figure*}[t]
\centering{
\includegraphics[width=0.90\textwidth]{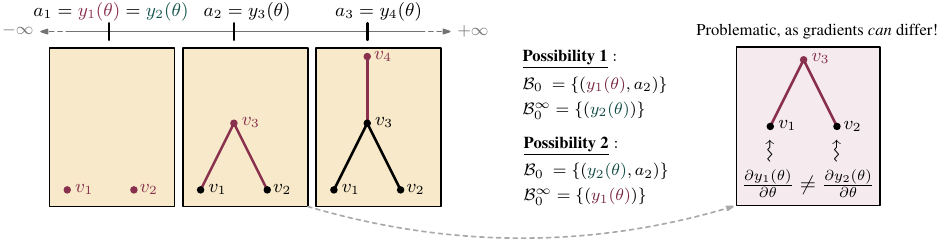}}
\caption{Illustration of a problematic scenario that can occur when
trying to backpropagate through the persistent homology computation.
The \emph{left-hand} side shows three steps in a filtration sequence 
with filtration values $a_i = y_i(\theta) = f(\theta,K,v_i)$, including the
two valid choices of the $0$-dimensional barcode $\mcB_0$ and $\mcB_0^\infty$ 
(essential). Numerically, the barcodes for both choices are equal, 
however, depending on the tie settling strategy, the gradients may differ. 
\label{fig:diff_problem}}
\end{figure*}

\textbf{Analysis of Lemma~\ref{lemma:filter_differentiable}.} A crucial assumption in Lemma~\ref{lemma:filter_differentiable} is that the 
filtration values are pairwise distinct. 
If this is not the case, the sorting permutation $\pi$ is not uniquely defined and Eq.~\eqref{lemma:filter_differentiable:eqn:filt} does not hold. 
Although the gradient \wrt~$\theta$ is still \emph{computable} in this situation, it depends on the particular implementation of the persistent homology algorithm. 
The reason for this is that the latter depends on a strict total ordering of the simplices such that
$\tau < \sigma$ whenever $\tau$ is a face of $\sigma$, formally $\tau \in \{\rho \subseteq \sigma : \dim \rho = \dim \sigma -1\}\Rightarrow \tau < \sigma$. %\in \partial_{\dim \sigma} (\sigma) .  
Specifically, the sublevel set filtration $(K^{f, i})_{i=1}^m$, only yields a \emph{partial} strict ordering by
\begin{align}
\label{eq:strict_order_1}
\sigma \in K_i, \tau \in K_j 	 \quad\text{and}\quad i<j &\Rightarrow \sigma < \tau\enspace, \\
\label{eq:strict_order_2}
\sigma, \tau \in K_i 			 \quad\text{and}\quad \dim(\sigma) < \dim(\tau)&\Rightarrow \sigma < \tau\enspace.
\end{align}
However, in the case 
$$\sigma, \tau \in K_i \land \dim(\sigma)=\dim(\tau)$$ 
we can neither infer $\sigma < \tau$ nor $\sigma > \tau$ from Eq. \eqref{eq:strict_order_1} and Eq.~\eqref{eq:strict_order_2}.
 Hence, those ``ties'' need to be settled in the implementation.  
If we were \emph{only} interested in the barcodes, the tie settling strategy 
would be irrelevant as the barcodes do not depend on the implementation. 
To see this, consider a point $(a,b)$ in 
the barcode $\mcB$. Now let
\begin{align*}
I_a & = \{i: 1 \leq {} i \leq n, f(\theta,K,v_i) = a\}, \text{ and} \\
I_b & = \{i: 1 \leq i \leq n, f(\theta,K,v_i) = b\}\enspace.
\end{align*}
Then, if $|I_a| > 1$ or $|I_b|>1$ the representation of 
$(a, b) \in \mcB$ may not be unique but dependent on the tie settling strategy used in the implementation of persistent homology.
In this case, the particular choice of tie settling strategy, could affect the gradient.
In Fig.~\ref{fig:diff_problem}, we illustrate this issue on a toy example of a problematic configuration.
Conclusively, different implementations yield the same barcodes, but probably different gradients. 
% $$
% (a, b) = \big(f(\theta,K,v_i),f(\theta,K,v_j)\big)~~~\text{with}~~~i \in I_a, j \in I_b
% $$
% are all valid representations of $(a,b)$ in $\mcB$. However, if the representation 
% is dependent on the tie settling strategy used in the persistent homology computation,
% the gradients may differ.
%
%
%
%but the representation of the values $a$ resp. $b$ is dependent on the selection $i\in I_a$ and $j \in I_b$.
%The selection, in turn, is determined by the tie settling strategy. 

%\tbd{Add noise references}
%Our pragmatic approach to address this issue is to settle ties randomly
%to obtain a strict total ordering. From a 
%theoretical perspective, this amounts to adding a sufficiently small amount of 
%noise $\varepsilon$ from a non-atomic distribution $\mathcal{D}$ with 
%an expected value of zero. In particular, we substitute
%$y_{i_j}(\theta)$ with $y_{i_j}(\theta)+\varepsilon_j$ for all
%$y_{i_1}(\theta) = y_{i_2}(\theta) \cdots = y_{i_q}(\theta)$ where
%$\varepsilon_i$ sampled i.i.d. from $\mathcal{D}$ to define the ordering.
%% 
\begin{rem}
An alternative, but computationally more expensive, strategy would be 
as follows: for a particular filtration value $a$, let $I_a = \{i: 1 \leq 
i \leq n, f(\theta,K,v_i) = a\}$. Now consider a point $(a,b)$ in the barcode with a ``problematic'' configuration, cf.~Fig.~\ref{fig:diff_problem} in $a$.
% where, say, in $a$ gradient computation is ``problematic'', \ie, implementation dependent. 
Upon setting
$$
(a,b) = \big(\nicefrac{1}{|I_a|} \sum_{i \in I_a} f(\theta,K,v_i),b\big)\enspace, 
$$
i.e., the (mean) aggregation of \emph{all possible} representations of $(a,b)$, 
this results in a gradient which is independent of the actual tie settling strategy. Yet, this is far more expensive to compute, due to the index
set construction of $I_a$, especially if $n$ is large. While 
it can be argued that this strategy is more ``natural``, it would lead
to a more involved proof for differentiability, which we leave for
future work.
\end{rem}

The difficulty of assigning/defining a proper gradient if the filtration values are 
\emph{not} pairwise distinct is not unique to our approach. In fact, other set 
operations frequently used in neural networks face a similar problem.
For example, consider the popular \texttt{maxpool} operator 
$\{y_1, \dots, y_n\} \mapsto \max(\{y_1, \dots, y_n\}) = z$, 
which is a well defined mapping. 
However, in the case of, say, $z = y_1 = y_2$, it is unclear if 
$y_1$ or $y_2$ is used to \emph{represent} its value $z$. 
In the situation of $z = y_1 = \varphi_{\theta}(x_1) = y_2 = \varphi_{\theta}(x_2)$ for some differentiable function $\varphi_{\theta}$, with differing gradients (\wrt~$\theta$) at $x_1$ and $x_2$ this could be problematic. 
In fact, the gradient \wrt~$\theta$ would then depend on the particular choice of representing $z$, \ie, the particular  \emph{implementation} of the \texttt{maxpool} operator. 

\subsection{Graph filtration learning (GFL)}

Having established the theoretical foundation of our approach, we now describe its practical application to graphs. 
First note that, as briefly discussed in 
\S\ref{section:background}, graphs \emph{are} simplicial complexes, 
although they are notationally represented in slightly different ways. 
For a graph $G=(V, E)$ we can directly define its simplicial complex by 
$K_G = \big\{\{v\}: v \in V \big\} \cup E$. 
We ignore this notational nuance and use $G$ and $K_G$ interchangeably. 
In fact, learning filtrations on graph-structured data integrates seamlessly into the presented framework. Specifically, the learnable vertex filter function, generically introduced in Definition \ref{defi:learnable_filter}, can be easily implemented by a neural network. If local node 
neighborhood information should be taken into account, this can be realized via a graph neural
network (\texttt{GNN}; see \citet{Wu19} for a comprehensive survey), operating on an initial node representation $l: \bbV \to \R^n$. The 
learnable vertex filter function then is a mapping of the form $v \mapsto \texttt{GNN}(G,l(v))$.

{\bf Selection of} $\mcV$. We use a vectorization based on a local 
weighting function, $s:\R^2\rightarrow\R$, of points in a given barcode, $
\mcB$. In particular, we select the \emph{rational hat} structure element 
of \cite{Hofer19b}, which is of the form
\begin{equation}
\label{eq:rat_hat}
  \mcB \ni p \mapsto \frac{1}{1+ \| p - c\|_1} - \frac{1}{1 + \big||r|-\|p - c \|_1\big|}
  \enspace, 
\end{equation}
where $c \in \R^2, r \in \R$ are learnable parameters. Intuitively, this 
can be seen as a function that evaluates the ``centrality'' of each point
$p$ of the barcode $\mcB$ \wrt~ a learnt center $c$ and a learnt shift/radius
$r$. The reason we prefer this variant over alternatives is that it yields a \emph{rational} 
dependency \wrt~$c$ and $r$, resulting in tamer gradient behavior during optimization. 
This is even more critical as, different to \cite{Hofer19b}, we optimize over parts of the model 
which appear before $\mcV$ and are thus directly dependent on its gradient.

We remark that our approach is not specifically bound to the choice in Eq.~\eqref{eq:rat_hat}. 
In fact, one could possibly use the strategies proposed by \citet{Carriere19a} or 
\citet{Zhao19} and we do not claim that our choice is optimal in any 
sense.

%\tbd{discuss: substituting differentiablity with sub-differentiablility also works for main theorem. $s$ is not %differentiable in some points but sub-differentiable}

% auto-ignore
% !TEX root = ./graph_filtration_learning.tex

\begin{table*}[t]
\small
\centering
\begin{tabu}{lcccc}
\toprule
\textbf{Method} & \texttt{REDDIT-BINARY} & \texttt{REDDIT-MULTI-5K} & \texttt{IMDB-BINARY} & \texttt{IMDB-MULTI} \\
\midrule
& \multicolumn{2}{l}{\emph{Initial node features: uninformative}} &\multicolumn{2}{l}{\emph{Initial node features: $l(v) = \deg(v)$}} \\
\midrule
\texttt{PH-only} 	    						& $90.3_{\pm 2.6}$ & $55.7_{\pm 2.1}$ & $68.9_{\pm 3.5}$ & $46.1_{\pm 4.2}$ \\ 
\texttt{1-GIN (GFL)}    									& $90.2_{\pm 2.8}$ & $55.7_{\pm 2.9}$ & $74.5_{\pm 4.6}$ & $49.7_{\pm 2.9}$ \\ 
\texttt{1-GIN (SUM)} \cite{Xu19a}    						& $81.2_{\pm 5.4}$ & $51.0_{\pm 2.2}$ & $73.5_{\pm 3.8}$ & $50.3_{\pm 2.6}$ \\
\texttt{1-GIN (SP)} \cite{Zhang18b}    			& $76.8_{\pm 3.6}$ & $48.5_{\pm 1.8}$ & $73.0_{\pm 4.0}$ & $50.5_{\pm 2.1}$ \\
\texttt{Baseline} \cite{Zaheer2017a} 			& $77.5_{\pm 4.2}$ & $45.7_{\pm 1.4}$ & $72.7_{\pm 4.6}$ & $49.9_{\pm 4.0}$ \\ 
\midrule
\multicolumn{3}{l}{\emph{State-of-the-Art (NN)}} \\
\midrule
\rowfont{\color{aurometalsaurus}}
\texttt{DCNN} \cite{Wang18a}				& n/a & n/a & $49.1$ & $33.5$ \\
\rowfont{\color{aurometalsaurus}}
\texttt{PatchySAN} \cite{Niepert16a} 	& $86.3$ & $49.1$ & $71.0$ & $45.2$ \\
\rowfont{\color{aurometalsaurus}}
\texttt{DGCNN} \cite{Zhang18b} 			& n/a & n/a & $70.0$ & $47.8$ \\
\rowfont{\color{aurometalsaurus}}
\texttt{1-2-3-GNN} \cite{Morris19a}		& n/a & n/a & $74.2$ & $49.5$ \\
\rowfont{\color{aurometalsaurus}}
\texttt{5-GIN (SUM)} \cite{Xu19a} & $88.9$ & $54.0$ & $74.0$ & $48.8$ \\ 
\bottomrule
\end{tabu} 
\caption{\label{table:social} Graph classification accuracies (with std. dev.), averaged
over ten cross-validation folds, on social network datasets. All 
\texttt{GIN} variants (including \texttt{5-GIN (Sum)}) were evaluated on exactly the same folds. The bottom part of 
the table lists results obtained by approaches from the literature. Operations in parentheses refer to the 
readout variant. Only \texttt{PH-only} \emph{always} uses the node degree.}
\end{table*}

\section{Experiments}
\label{section:experiments}

We evaluate the utility of the proposed homological
readout operation \wrt~different aspects. \emph{First}, as argued 
in \S\ref{section:introduction}, we want to avoid the challenge of 
tuning the number of used message passing iterations.  
To this end, when applying our readout operation, referred to as 
\texttt{GFL}, we learn the filter function from just one round 
of message passing, i.e., the most local, non-trivial variant.  
\emph{Second}, we aim to address the question of whether learning 
a filter function is actually beneficial, compared to defining the filter function 
a-priori, to which we refer to as \texttt{PH-only}.

Importantly, to clearly assess the power of a readout operation 
across various datasets, we need to first investigate whether 
the  discriminative power of a representation is not primarily 
contained in the initial node features. In fact, if a baseline 
approach using only node features performs en par with approaches 
that leverage the graph structure, this would indicate that detailed
connectivity information is of little relevance for the task. 
In this case, 
we cannot expect that a readout which strongly depends on the 
latter is beneficial.

\textbf{Datasets.} We use two common benchmark datasets for graphs with
discrete node attributes, i.e., \texttt{PROTEINS} and \texttt{NCI1}, 
as well as four social network datasets (\texttt{IMDB-BINARY}, \texttt{IMDB-MULTI}, 
\texttt{REDDIT-BINARY}, \texttt{REDDIT-5k}) which \emph{do not} contain any 
node attributes (see supplementary material).

\subsection{Implementation}

We provide a high-level description of the 
implementation\footnote{Source code is publicly available at \url{https://github.com/c-hofer/graph_filtration_learning}.} (cf. Fig.~\ref{fig:intro}) and refer the reader to 
the supplementary material for technical details. 

First, to implement 
the filter function $f$, we use a single GIN-$\varepsilon$ layer of \cite{Xu19a} for 
one level of message passing (i.e., \texttt{1-GIN}) with hidden dimensionality of $64$. The 
obtained latent node representation is then passed through a two 
layer MLP, mapping from $\mathbb{R}^{64}\to[0,1]$, with a sigmoid 
activation at the last layer. Node degrees and (if available) 
discrete node attributes are encoded via embedding layers with 
64 dimensions. In case of \texttt{REDDIT-*} graphs, initial
node features are set \emph{uninformative}, i.e., vectors of all ones 
(following the setup by \citet{Xu19a}).

Using the output of the vertex filter, persistent homology is computed via a 
parallel GPU variant of the original reduction algorithm from \cite{Edelsbrunner02a}, 
implemented in \texttt{PyTorch}. In particular, we first compute $0$- and 
$1$-dimensional barcodes for $f$ and $g=-f$, which correspond to
sub- and super levelset filtrations. For each filter function, i.e.,
$f$ and $g$, we obtain barcodes for $0$- and $1$-dim. essential
features, as well as $0$-dim. non-essential features.
Non-essential points in barcodes \wrt~$g$ are mapped into $[0,1]^2$ 
by mirroring along the main diagonals and essential points 
(i.e., birth-times) are mapped to $[0,1]$ by mirroring around $0$. 
We then take the union of the barcodes corresponding to sub- and 
super levelset filtrations which reduces the number of processed barcodes from 6 to 3 (see Fig.~\ref{fig:intro}).

Finally, each barcode is passed through a vectorization layer
implementing the barcode coordinate function $\mcV$ of Definition~\ref{defi:barcode_coordinate_function} 
using Eq.~\eqref{eq:rat_hat}. We use $100$ output dimensions for 
each barcode. Upon concatenation, this results in a $300$-dim. representation of a
graph (\ie, the output of our readout) that is passed to a final 
\texttt{MLP}, implementing the classifier. 

\textbf{Baseline and readout operations.} The previously mentioned \texttt{Baseline}, agnostic
to the graph structure, is implemented using the deep sets 
approach of \cite{Zaheer2017a}, which is similar to the \texttt{GIN}
architecture used in all experiments, without any message passing. 
In terms of readout functions, we compare against the prevalent 
sum aggregation (\texttt{SUM}), as well as sort pooling (\texttt{SP}) from \cite{Zhang18b}. 
We do not explicitly compare against differentiable pooling from 
\cite{Ying18a}, since we allow just one level of message passing, 
rendering differentiable pooling equivalent to \texttt{1-GIN (SUM)}.

\textbf{Training and evaluation.} We train for $100$ epochs using ADAM with
an initial learning rate of $0.01$ (halved every $20$-th epoch) and a weight
decay of $10^{-6}$. No hyperparameter tuning or early stopping 
is used. For evaluation, we follow previous work \citep[see, e.g.,][]{Morris19a,Zhang18b}
and report cross-validation accuracy, averaged over ten folds, of the 
model obtained in the final training epoch.

\subsection{Complexity \& Runtime}

For the standard persistent homology matrix reduction algorithm \cite{Edelsbrunner02a}, the computational complexity is, in the worst case, $\mathcal{O}(m^3)$ where $m$ denotes the number of simplices. However, as remarked in the literature \citep[see][]{Bauer17a}, computation scales quasi-linear in practice. In fact, it has been shown that persistent homology can be computed in matrix multiplication time \cite{Milosavljevic11a}, \eg, in $\mathcal{O}(m^{2.376})$
using Coppersmith-Winograd. For $0$-dimensional persistent homology, union-find data structures can be used, reducing complexity to $\mathcal{O}(m\alpha^{-1}(m))$, where $\alpha^{-1}(\cdot)$ denotes the (slowly-growing) inverse of the Ackermann function. 

\begin{figure}
\begin{center}
\includegraphics[width=0.95\columnwidth]{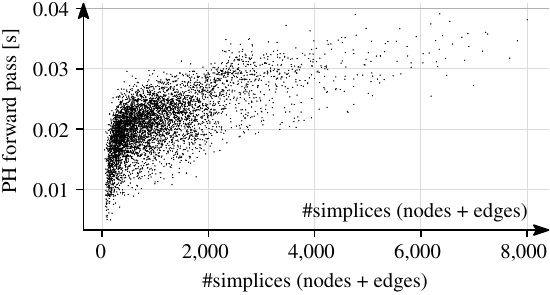}
\end{center}
\vspace{-4pt}
\caption{\label{fig:timing} \texttt{REDDIT-MULTI-5K} timing experiments 
(in [s]) for persistent homology computation during one forward 
pass.}
\vspace{-5pt}
\end{figure}

In our work, we rely on a \texttt{PyTorch}-compatible, (na\"ive) parallel GPU 
variant of the original matrix reduction algorithm for persistent homology (adapted from \cite{Hofer19a}). For runtime assessment, we consider the proposed readout operation and measure its runtime during a forward pass (per graph). Fig.~\ref{fig:timing} shows that runtime  indeed scales quasi-linearly with the number of simplices (\ie, vertices and edges) per graph. The average runtime (on \texttt{Reddit-5K}) is $0.02$ [s] vs. $0.002$ [s] (4-GIN, factor $\approx 10\times$) and $0.0004$ [s] (1-GIN, factor $\approx 50\times $). We compare to a 4-GIN variant for a fair comparison (as our readout also needs one level of message passing).
However, these results have to be interpreted with caution, as GPU-based persistent homology computation is still in its infancy and leaves substantial room for improvements.
(\eg, computing cohomology instead, or implementing optimization as in \citet{Bauer14b}).

\subsection{Results}
\label{subsection:results}

Table~\ref{table:social} lists the results for the social networks. On both 
\texttt{IMDB} datasets, the \texttt{Baseline} performs en par with the state-of-the-art, 
as well as all readout strategies. However, this is not surprising as the graphs are 
very densely connected and substantial information is already encoded in the node degree. Consequently, 
using the degree as an initial node feature, combined
with GNNs and various forms of readout does not lead to noticeable improvements.

\texttt{REDDIT-*} graphs, on the other hand, are far from fully connected and 
the global structure is more hierarchical. Here, the information captured by
persistent homology is highly discriminative, with GFL outperforming \texttt{SUM} and 
\texttt{SP} (sort pooling) by a large margin. 
Notably, setting the filter function a-priori to the degree function performs equally well. 
We argue that this might already be an optimal choice on \texttt{REDDIT}. On other
datasets (e.g., \texttt{IMDB}) this is not the case. Importantly, 
although the initial features on \texttt{REDDIT} graphs are set uninformative 
for all readout variants with \texttt{1-GIN}, \texttt{GFL} can learn a filter
function that is equally informative.

% \begin{wraptable}{1}{70mm}
\begin{table}[t]
\small
\begin{tabular}{lcc}
\toprule
\textbf{Method} & \texttt{PROTEINS} & \texttt{NCI1} \\
\midrule
\multicolumn{3}{l}{\emph{Initial node features: $l(v) = \deg(v)$}} \\
\midrule
\texttt{PH-only} 	    							& $73.5_{\pm 3.0}$ & $67.8_{\pm 2.2}$ \\ 
\texttt{1-GIN (GFL)}    							& $74.1_{\pm 3.4}$ & $71.2_{\pm 2.1}$ \\ 
\texttt{1-GIN (SUM)} \cite{Xu19a}    						& $72.1_{\pm 3.5}$ & $67.4_{\pm 3.0}$\\
\texttt{1-GIN (SP)} \cite{Zhang18b}    				& $72.4_{\pm 3.1}$ & $67.8_{\pm 2.2}$\\
\texttt{Baseline} \cite{Zaheer2017a}						& $73.1_{\pm 3.7}$ & $65.8_{\pm 2.8}$ \\
\midrule
\multicolumn{3}{l}{\emph{Initial node features $l(v) = [\deg(v),\lab(v)]$}} \\
\midrule
\texttt{PH-only} 	    							& n/a & n/a \\
\texttt{1-GIN (GFL)}    							& $73.4_{\pm 2.9}$ & $77.2_{\pm 2.6}$ \\ 
\texttt{1-GIN (SUM)} \cite{Xu19a}    						& $75.1_{\pm 2.8}$ & $77.4_{\pm 2.1}$ \\ 
\texttt{1-GIN (SP)} \cite{Zhang18b}   			& $73.5_{\pm 3.8}$ & $76.9_{\pm 2.3}$ \\ 
\texttt{Baseline} \cite{Zaheer2017a} 			& $73.8_{\pm 4.4}$ & $67.2_{\pm 2.3}$\\
\midrule
\multicolumn{3}{l}{\emph{State-of-the-Art (NN)}} \\
\midrule
\textcolor{aurometalsaurus}{\texttt{DCNN}}      \cite{Wang18a}			& \textcolor{aurometalsaurus}{$61.3$} & \textcolor{aurometalsaurus}{$62.6$} \\
\textcolor{aurometalsaurus}{\texttt{PatchySAN}}  \cite{Niepert16a}  		& \textcolor{aurometalsaurus}{$75.9$} & \textcolor{aurometalsaurus}{$78.6$} \\
\textcolor{aurometalsaurus}{\texttt{DGCNN}} 		\cite{Zhang18b}     & \textcolor{aurometalsaurus}{$75.5$} & \textcolor{aurometalsaurus}{$74.4$} \\
\textcolor{aurometalsaurus}{\texttt{1-2-3-GNN}} \cite{Morris19a}      & \textcolor{aurometalsaurus}{$75.5$} & 
\textcolor{aurometalsaurus}{$76.2$} \\
\textcolor{aurometalsaurus}{\texttt{5-GIN (SUM)}} \cite{Xu19a}       & \textcolor{aurometalsaurus}{$71.2$} & 
\textcolor{aurometalsaurus}{$75.9$} \\
\bottomrule
\end{tabular}
\caption{\label{table:bio}  Graph classification accuracies (with std. dev.), averaged
over ten cross-validation folds, on graphs \emph{with} node attributes. Comparison to 
the state-of-the-art follows the results of Table~\ref{table:social}.}
\vspace{-5pt}
\end{table}
% \end{wraptable}

Table~\ref{table:bio} lists results on \texttt{NCI1} and \texttt{PROTEINS}.
On the latter, we observe that already the \texttt{Baseline}, agnostic to the connectivity information, 
is competitive to the state-of-the-art. GNNs with different readout strategies, including ours, 
only marginally improve performance. It is therefore challenging, to assess the utility of different
readout variants on this dataset. On \texttt{NCI1}, the 
situation is different. Relying on node degrees only, our GFL readout clearly outperforms the 
other readout strategies. This indicates that \texttt{GFL} can successfully capture  
the underlying discriminative graph structure, relying only on minimal information gathered
at node-level. Including label information leads to results competitive to the 
state-of-the-art without explicitly tuning the architecture. While all other 
readout strategies equally benefit from additional node attributes, the fact that 
\texttt{5-GIN (SUM)} performs worse than \texttt{1-GIN (SUM)} highlights our argument 
that message passing needs careful architectural design.

% auto-ignore
% !TeX root = ./graph_filtration_learning.tex

\section{Discussion}
\label{section:discussion}

To the best of our knowledge, we introduced the first approach to actively integrate 
persistent homology within the realm of GNNs, 
offering GFL as a novel type of readout.
As demonstrated throughout all experiments, GFL,  
which is based on the idea of filtration learning, is able to achieve 
results competitive to the state-of-the-art on various datasets. 
Most notably, this is achieved with a single architecture that 
only relies on a simple one-level message passing scheme. 
This is different to previous works, where the amount of information that 
is iteratively aggregated via message passing can be crucial. We also 
highlight that \texttt{GFL} could be easily extended to incorporate 
edge level information, or be directly used on graphs with continuous 
node attributes. 

As for the additional value of our readout operation, 
% we argue that 
it leverages the global graph structure for aggregating the local node-based information.
% in a theoretically well-founded manner. 
%In other words, local information 
%is inherently coupled to the overall structure through the readout operation. 
Theoretically, we established how 
to backpropagate a (gradient-based) learning signal through the persistent 
homology computation in combination with a differentiable vectorization 
scheme. For future work, it is interesting to study 
additional filtration techniques (\eg, based on edges, or cliques) and 
whether this is beneficial.

\section*{Acknowledgements}
This work was partially funded by the Austrian Science Fund 
(FWF): project FWF P31799-N38 and the Land Salzburg (WISS 2025) under project numbers
20102-F1901166-KZP and 20204-WISS/225/197-2019.

% \section*{References}

\bibliography{booksLibrary,libraryChecked,libraryBR}
\bibliographystyle{icml2020}

\end{document}

\onecolumn
\icmltitle{Graph Filtration Learning -- Supplementary Material}

This supplementary material contains the full proof of Lemma~1 omitted in the main work and additional information to the used datasets. It further contains details to the implementation of the models used in the experiments. For readability, all necessary definitions and results are restated and the numbering matches 
the original numbering.

\section{Dataset details}

The following table contains a selection of statistics relevant to the datasets used in our experiments.

%\begin{table}[h!]
%\caption{
%Meta data of graph datasets.
%\label{tbl:results_main}}
\begin{small}
\begin{center}
\begin{tabular}{lcccccc}
\toprule
Datasets &  \footnotesize{\texttt{REDDIT-BINARY}} & \footnotesize{\texttt{REDDIT-MULTI-5K}} & \footnotesize{\texttt{IMDB-BINARY}} & \footnotesize{\texttt{IMDB-MULTI}} & \footnotesize{\texttt{PROTEINS}} & \footnotesize{\texttt{NCI1}}\\
\midrule
\# graphs  & $2000$ & $4999$ & $1000$ & $1500$ & $1113$ & $4110$ \\
\# classes  & $2$ & $5$ & $2$ & $3$ & $2$ & $2$ \\
$\varnothing$ nodes  & $429.6$ & $508.5$ & $19.8$ & $13.0$ & $39.1$ & $29.9$ \\
$\varnothing$ edges  & $497.8$ & $594.9$ & $96.5$ & $65.9$ & $72.8$ & $32.3$ \\
\# labels & n/a& n/a& n/a& n/a& $3$& $37$\\
\bottomrule
\end{tabular}
\end{center}
\end{small}

\section{Proof of Lemma~1}

% auto-ignore
% !TEX root = ./graph_filtration_learning.tex

\vskip1ex
\begin{defi}[Learnable vertex filter function]
\label{defi:learnable_filter}
Let $\bbV$ be a vertex domain, $\bbK$ the set of possible simplicial complexes over $\bbV$
and let
\[
f: \R \times \bbK \times \bbV \rightarrow \R \quad\quad (\theta, K, v) \mapsto f(\theta,K, v)
\]
be differentiable in $\theta$ for $K \in \bbK, v \in \bbV$. Then, we call $f$ a \emph{learnable vertex filter function} with parameter $\theta$. 
\end{defi}

% auto-ignore
% !TEX root = ./graph_filtration_learning.tex

\begin{defi}[Barcode coordinate function]
\label{defi:barcode_coordinate_function}
Let $s: \R^2 \rightarrow \R$ be a differentiable function that vanishes on the
diagonal of $\mathbb{R}^2$. Then 
\[
  \mcV: \bbB \rightarrow \R \quad \quad \mcB \mapsto \sum\limits_{(b, d) \in \mcB}s(b, d)
\]
is called \emph{barcode coordinate function}. 
\label{defi:barcode_coordinate_function}
\end{defi}

% auto-ignore
% !TEX root = ./graph_filtration_learning.tex

\begin{lem}
\label{lemma:filter_differentiable}
Let $K$ be  a \emph{finite} simplicial complex with vertex set $V=\{v_1, \dots, v_n\}$, $f: \R \times \bbK \times \bbV \rightarrow \R$ be a learnable vertex filter function as in Definition \ref{defi:learnable_filter} and $\mcV$ a barcode coordinate function as in Definition \ref{defi:barcode_coordinate_function}.
If, for $\theta_0 \in \R$, it holds that the pairwise vertex filter values are distinct, \ie, 
\[
  f(\theta_0, K, v_i) \neq f(\theta_0,K,  v_j) 
  \quad \text{ for } \quad 
  1 \leq i < j \leq n
\]
then the mapping
\begin{equation}
  \theta \mapsto \mcV\big(\ph_k^{f}(\theta, K)\big)
\label{eqn:lemma:pairwise_filter_values_differentiable}
\end{equation}
is \emph{differentiable} at $\theta_0$. 
\end{lem}
\begin{proof}
For notational convenience, let $y_i = f(\theta_0, K, v_i) = f(\theta_0,v_i)$.
Also, let $\pi$ the sorting permutation of 
$\big(y_i(\theta_0)\big)_{i=1}^n$, \ie, 
$y_{\pi(1)}(\theta_0) < y_{\pi(2)}(\theta_0) < \dots < y_{\pi(n)}(\theta_0)$. 
By assumption  the pairwise filter values are distinct, thus there is a neighborhood around $\theta_0$ such that the ordering of the filtration values is not modified by changes of $\theta$ within this neighborhood, \ie, 
\begin{equation}
\label{eq:lemma:filter_diffable_important_step_1}
  \exists \varepsilon > 0 \forall h \in \R: |h| < \varepsilon
  \Rightarrow
  y_i(\theta_0+h) \neq y_j(\theta_0+h)
\end{equation}
and
\begin{equation}
\label{eq:lemma:filter_diffable_important_step_2}
  y_{\pi(1)}(\theta_0+h) < y_{\pi(2)}(\theta_0+h) < \dots < y_{\pi(n)}(\theta_0+h)
  \enspace.  
\end{equation}
This implies that a sufficiently small change, $h$, of $\theta_0$ does not change the 
induced filtrations. 
Formally,
\begin{equation}
\big(K_i^f(\theta_0)\big)_{i=0}^n = \big(K_i^f(\theta_0 + h)\big)_{i=0}^n
\quad
\text{with}
\quad
0 \leq i \leq n
\enspace. 
\label{eqn:filtration_equality}
\end{equation}
Importantly, this means that
\begin{equation}
\label{eqn:mu_equality}
  \mu_k^{i, j}(\theta_0) =  \mu_k^{i, j}(\theta_0 + h)
  \quad
  \text{for}
  \quad
  1 \leq i < j \leq n
  \enspace. 
\end{equation}

We next show that the derivative of Eq.~\eqref{eqn:lemma:pairwise_filter_values_differentiable} with 
respect to $h$ exists. By assumption, $s$ is differentiable and thus 
$s\big(f(\cdot, v_i), s(f(\cdot, v_j)\big)$ is differentiable. Now consider
\begin{small}
% \begin{equation*} 
% \begin{split}
\begin{align*}
  &\lim\limits_{|h| \rightarrow 0}   
  \frac{\mcV(\ph_k^f(K,\theta_0)) - \mcV(\ph_k^f(K,\theta_0+h))}{h}\\  
  &=
  \lim\limits_{|h| \rightarrow 0}   
  \frac{\sum\limits_{i < j} \mu_k^{i, j}(\theta_0) \cdot s\big(y_{\pi(i)}(\theta_0), y_{\pi(j)}(\theta_0)\big) - \sum\limits_{i < j} \mu_k^{i, j}(\theta_0 + h) \cdot s\big(y_{\pi(i)}(\theta_0+h), y_{\pi(j)}(\theta_0+h)\big)}
  {h}
  \\
  &=
    \lim\limits_{|h| \rightarrow 0}   
  \frac{\sum\limits_{i < j}\mu_k^{i, j}(\theta_0)\cdot\left[s\big(y_{\pi(i)}(\theta_0), y_{\pi(j)}(\theta_0)\big)-s\big(y_{\pi(i)}(\theta_0+h), y_{\pi(j)}(\theta_0+h)\big)\right]}
  {h}=
  \tag{ by Eq.~\eqref{eqn:mu_equality}}
  \\
  &=
  \sum\limits_{i < j}
  \mu_k^{i, j}(\theta_0) \cdot 
  \lim\limits_{|h| \rightarrow 0}
  % \left[
  \frac{s\big(y_{\pi(i)}(\theta_0), y_{\pi(j)}(\theta_0)\big)-s\big(y_{\pi(i)}(\theta_0+h), y_{\pi(j)}(\theta_0+h)\big)}
  {h}
  \\
  &=
  \sum\limits_{i < j}
  \mu_k^{i, j}(\theta_0) \cdot 
  \lim\limits_{|h| \rightarrow 0}
  \frac{s\big(f(\theta_0, v_{\pi(i)}), f(\theta_0, v_{\pi(j)})\big)-s\big(f(\theta_0+h, v_{\pi(i)}), f(\theta_0+h, v_{\pi(j)})\big)}
  {h}
  \\
  &=
  \sum\limits_{i < j}
  \mu_k^{i, j}(\theta_0) \cdot 
  \frac{\partial s\big(f(\theta, v_{\pi(i)}\big), f\big(\theta, v_{\pi(j)})\big)}{\partial \theta}(\theta_0)
  \enspace.
\end{align*}
% \end{split}
% \end{equation*}
\end{small}
This concludes the proof, since the derivative within the summation exists by assumption. 
\end{proof}

\section{Architectural details}

As mentioned in Section~5 (Experiments), we implement the learnable vertex filter function $f(\theta,K,v_i)$
using a single \texttt{GIN}-$\varepsilon$ layer from \cite{Xu19a} (with $\varepsilon$ 
set as a learnable parameter). The internal architecture is as follows: 
\begin{center}
\texttt{Embedding[n,64]-FC[64,64]-BatchNorm-LeakyReLU-FC(64,64)}.
\end{center}
Here, \texttt{n} denotes the dimension of the node attributes. For example, if initial
node features are based on the degree function and the maximum degree over all graphs 
is 200, then \texttt{n=200+1}. In other words, \texttt{n} is the \emph{number of
embedding vectors} in $\mathbb{R}^{64}$ used to represent node degrees.

The 
multi-layer perceptron (\texttt{MLP}) mapping the output of the \texttt{GIN} layer
to a real-valued node filtration value is parameterized as:
\begin{center}
\texttt{Embedding[64,64]-FC[64,64]-BatchNorm-LeakyReLU-FC(64,1)-Sigmoid}.
\end{center}
As classifier, we use a simple \texttt{MLP} of the form 
\begin{center}
\texttt{FC[300,64]-ReLU-FC[64,\#classes]}.
\end{center}
Here, the input dimensionality is 300, as each barcode is represented by
a 100-dimensional vector.

% \section*{References}

\bibliography{booksLibrary,libraryChecked}
\bibliographystyle{icml2020}